\documentclass[10pt,twocolumn,letterpaper]{article}

\usepackage{arxiv}
\usepackage{times}
\usepackage{epstopdf}
\usepackage{graphicx}
\usepackage{amsmath}
\usepackage{amssymb}
\usepackage[titletoc,title]{appendix}

\usepackage{enumitem}
\usepackage{gensymb}
\usepackage[font=small,labelfont=bf]{caption}
\usepackage{subcaption}
\usepackage{multirow}
\DeclareMathOperator*{\argmax}{arg\,max}

\newcommand{\norm}[1]{\left\lVert#1\right\rVert}

\newcommand{\figref}[1]{Fig. \ref{#1}}
\newcommand{\tabref}[1]{Table \ref{#1}}
\renewcommand{\eqref}[1]{Eq. \ref{#1}}
\renewcommand{\paragraph}[1]{\smallskip{\bf{#1}\;}}
\graphicspath{ {./figures/} }

\usepackage[pagebackref=true,breaklinks=true,letterpaper=true,colorlinks,bookmarks=false]{hyperref}

\iccvfinalcopy 

\begin{document}

\title{ACE-Net: Fine-Level Face Alignment through Anchors and Contours Estimation}

\author{Jihua Huang\\
SRI International\\
201 Washington Rd, Princeton, NJ 08540\\
{\tt\small jihua.huang@sri.com}
\and
Amir Tamrakar\\
SRI International\\
201 Washington Rd, Princeton, NJ 08540\\
{\tt\small secondauthor@i2.org}
}

\maketitle

\begin{abstract}
We propose a novel facial \textbf{A}nchors and \textbf{C}ontours \textbf{E}stimation framework, \textbf{ACE-Net}, for fine-level face alignment tasks. ACE-Net predicts facial anchors and contours that are richer than traditional facial landmarks while overcoming ambiguities and inconsistencies in their definitions. We introduce a weakly supervised loss enabling ACE-Net to learn from existing facial landmarks datasets without the need for reannotation. Instead, synthetic data, from which GT contours can be easily obtained, is used during training to bridge the density gap between landmarks and true facial contours. We evaluate the face alignment accuracy of ACE-Net with respect to the HELEN dataset which has 194 annotated facial landmarks, while it is trained with only 68 or 36 landmarks from the 300-W dataset. We show that ACE-Net generated contours are better than contours interpolated straight from the 68 GT landmarks and ACE-Net also outperforms models trained only with full supervision from GT landmarks-based contours.
\end{abstract}

\section{Introduction}
\begin{figure}[!tbp]
    \centering
    \captionsetup[subfigure]{belowskip=-5pt}
    \begin{subfigure}[b]{0.23\textwidth}
        \includegraphics[trim=32 32 32 32, clip, width=\textwidth]{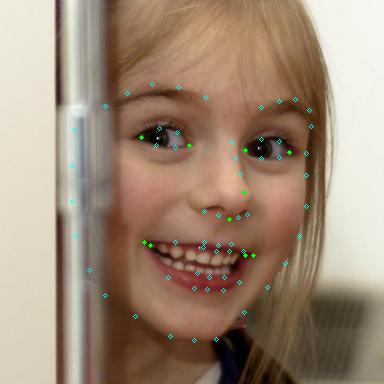}
        \caption{Landmarks representation}
        \label{fig:landmarks_vs_AC-landmarks}
    \end{subfigure}
    \hfill
    \begin{subfigure}[b]{0.23\textwidth}
        \includegraphics[trim=32 32 32 32, clip, width=\textwidth]{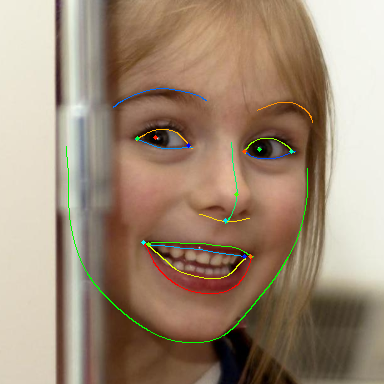}
        \caption{AC representation}
        \label{fig:landmarks_vs_AC-AC}
    \end{subfigure}
    \\[2ex]
    \begin{subfigure}[b]{0.153\textwidth}
        \includegraphics[trim=130 80 120 227, clip, width=\textwidth]{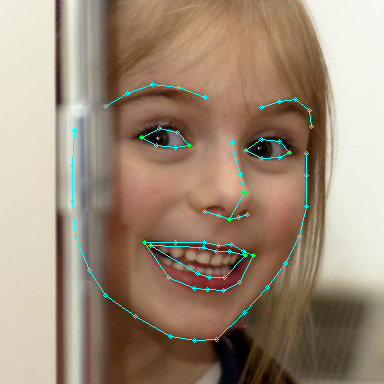}
        \caption{Line-contours}
        \label{fig:landmarks_vs_AC-line-zoom}
    \end{subfigure}
    \begin{subfigure}[b]{0.153\textwidth}
        \includegraphics[trim=130 80 120 227, clip, width=\textwidth]{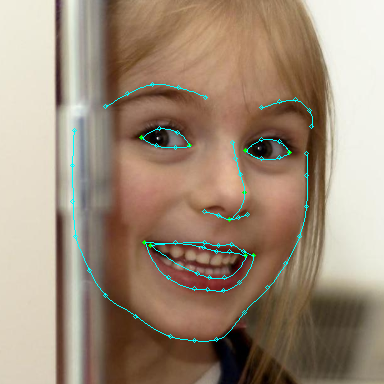}
        \caption{Interpolated}
        \label{fig:landmarks_vs_AC-interp-zoom}
    \end{subfigure}
    \begin{subfigure}[b]{0.153\textwidth}
        \includegraphics[trim=130 80 120 227, clip, width=\textwidth]{landmarks_vs_AC-AC.png}
        \caption{Predicted AC}
        \label{fig:landmarks_vs_AC-AC-zoom}
    \end{subfigure}
    \caption{(a) Facial landmarks showing the Anchor landmarks (filled) and Contour landmarks (unfilled). (b) The Anchor-Contour (AC) representation. Anchor landmarks are the same and full contours (colored curves) replace the contour landmarks. Zooming in: (c) contours generated by connecting the landmarks with line segments and (d) contours generated by interpolating the landmarks with splines are not as good as (e) contours from our ACE-Net. }
    \label{fig:landmarks_vs_AC}
\end{figure}

Face alignment is the basis for various types of face analysis such as face mesh reconstruction \cite{aldrian2012inverse, bas2016fitting, tuan2017regressing, sanyal2019learning, zhu2020reda}, facial behavior modeling for deepfake detection \cite{agarwal2019protecting} and visual speech recognition \cite{jha2018word, morrone2019face, jang2019lip}, where fine-level alignment/tracking is crucial. Most of the existing approaches use facial landmarks to perform face alignment. While the accuracy of facial landmarks detection has greatly improved over the past few years, there remain three major issues with facial landmarks representation: First, only a small subset of landmarks are well defined and can be localized accurately, such as eye corners. We refer to these as \emph{anchors}. The rest of the landmarks are just discrete points sampled along various facial contours. We call those \emph{contour landmarks} as their positions are ambiguous along the contours (\figref{fig:landmarks_vs_AC-landmarks}). Second, typical facial landmarks definitions are not dense enough to capture fine-level details of facial contours (Fig. \ref{fig:landmarks_vs_AC-line-zoom},\ref{fig:landmarks_vs_AC-interp-zoom}). Third, facial landmarks definitions are inconsistent across existing datasets \cite{koestinger2011annotated,sagonas2013300,wu2018look,le2012interactive}. 

\begin{figure}[!tbp]
    \captionsetup[subfigure]{belowskip=-5pt}
    \begin{subfigure}[b]{0.18\textwidth}
        \includegraphics[width=\textwidth]{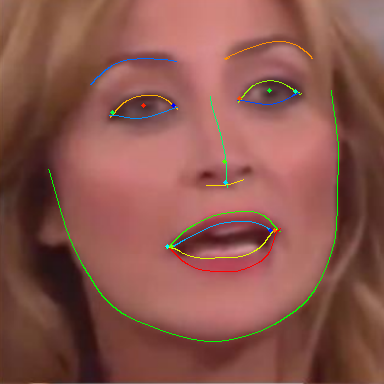}
        \caption{AC}
    \end{subfigure}
    \begin{subfigure}[b]{0.28\textwidth}
        \includegraphics[trim=0 5 0 5, clip, width=\textwidth]{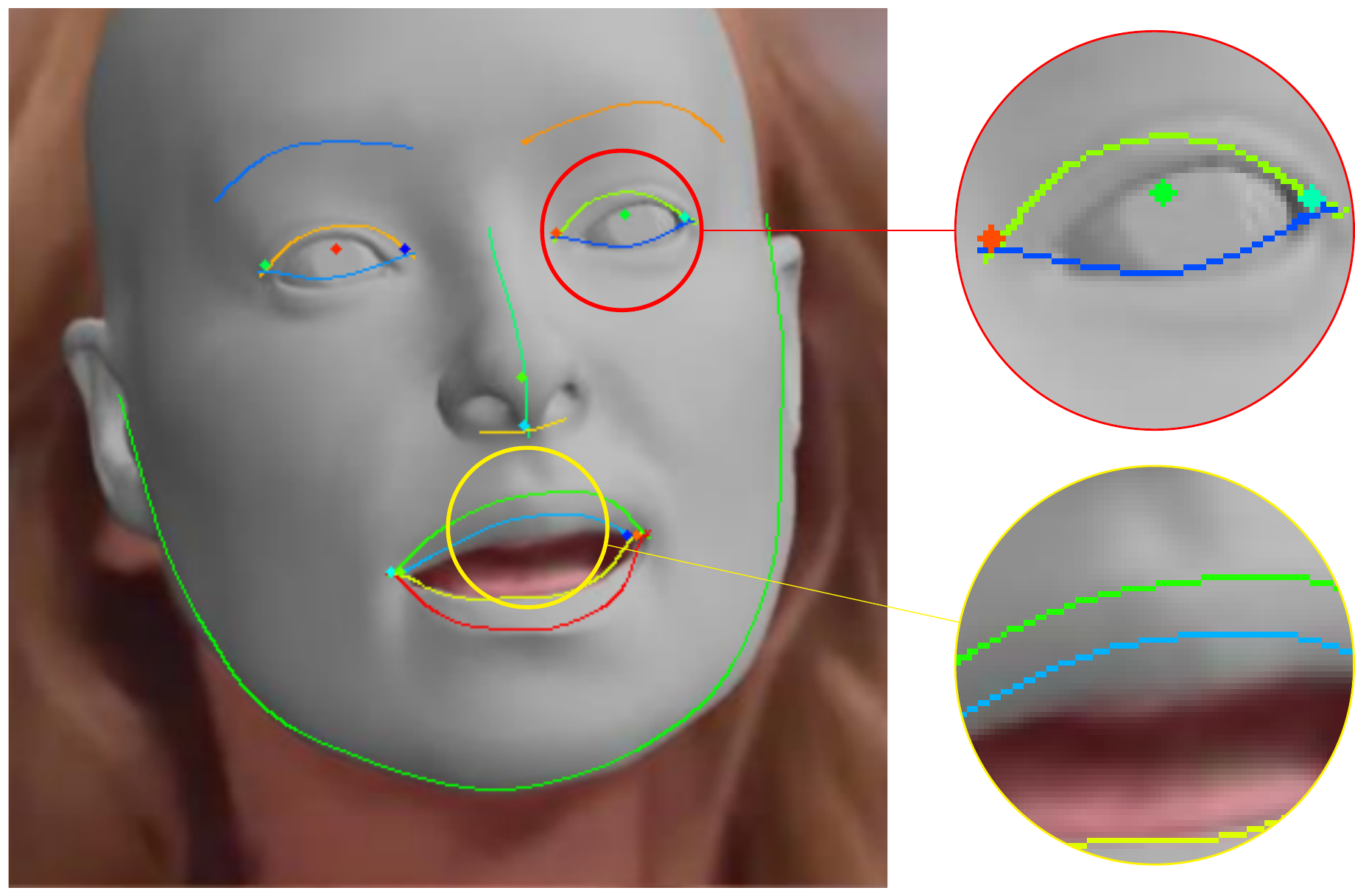}
        \caption{Face mesh}
    \end{subfigure}
    \caption{(a) AC representation extracted by ACE-Net. (b) Face Mesh from \cite{zhu2020reda} with AC overlaid show that there is room for improvement in the mesh fitting that AC rep can provide (see selected zoom-ins). For motivation only (not discussed in this paper.)     }
    \label{fig:mesh_vs_AC}
\end{figure}

In order to address these issues, we propose a new representation that retains all the anchor landmarks but replaces all the contour landmarks with actual contours. We call this the \textbf{Anchors and Contours (AC)} representation (\figref{fig:landmarks_vs_AC-AC}). These face contours provide better constraints for alignment, for example when fitting meshes (\figref{fig:mesh_vs_AC}). In this paper, we define a new framework which we call facial \textbf{Anchors and Contours Estimation Network (ACE-Net)} to extract the AC representations from images. ACE-Net consist of two modules, (i) the \emph{AC prediction module} which predicts AC heatmaps from input images and (ii) the \emph{AC extraction module} which converts the extracted AC heatmaps to the final AC representations with sub-pixel accuracy. 

A new representation typically necessitates reannotation of the datasets with the new representation to generate the necessary training data for the new model. However, facial contours are extremely time consuming to annotate and presents a formidable bottleneck. Instead, we devise two strategies that allow us to bypass this bottleneck. First, we introduce a novel weakly supervised loss that enforces the predicted AC to pass through existing annotated facial landmarks. This allows us to use existing facial landmarks datasets as is. An added benefit is that this makes us independent of the specific landmarks definitions and allows us to use a multitude of datasets. We show later that a 36-landmarks-based weak supervision  (vs the typical 68) still works quite well.  Second, since the density of contour landmarks are insufficient to fully characterize the contours, we augment the training with synthetic data that provides contours for full supervision. We find that in spite of the domain gap, synthetic data helps improve the accuracy of the contour localizations in the spaces between landmarks. Our ACE-Net is thus trained with weak supervision from the landmarks and full supervision from the synthetic contours. 

We evaluate ACE-Net on the densely annotated HELEN dataset \cite{le2012interactive}. We compare ACE-Net results against line-contours (generated by connecting landmarks with line segments) and interpolated-contours (generated by spline fitting through the landmarks) derived straight from the GT landmarks. Since GT landmarks are arguably better than any SOTA landmark detectors, these contours can be considered the best AC results possible directly from landmarks. Furthermore, instead of using full supervision from the synthetic data, we also experiment with using these GT landmarks-derived contours to provide the full supervision (in which case, there is no need for the weak loss). Both our qualitative and quantitative results show that at a fine-level, ACE-Net outperforms landmarks based models, especially in regions lacking training annotations. 

In summary, our main contributions include:
\begin{itemize}[topsep=3pt]
    \itemsep 0em
    \item ACE-Net, the first facial alignment framework that predicts facial anchors and contours (with sub-pixel accuracy) instead of just facial landmarks.
    \item A novel ``contourness" loss to effectively use existing facial landmarks as weak supervision. 
    \item The incorporation of synthetic data to complement the training by bridging the gap between landmarks and contours (despite the domain gap) thus bypassing the need for re-annotation. We show that this is better than using landmark-interpolated contours. 
    \item An evaluation metric for the AC representations with respect to densely annotated facial landmarks datasets.
\end{itemize}

\section{Related Work}
Classic face alignment approaches are mostly based on AAMs \cite{cootes2001active, saragih2007nonlinear, sauer2011accurate}, ASMs \cite{cootes1992active,cootes1995active}, CLMs \cite{cootes2012robust,saragih2011deformable} or Cascaded Regression \cite{feng2015cascaded,xiong2013supervised}. Recent advances mainly focus on two major categories of deep learning based methods: coordinate regression and heatmap regression.

\paragraph{Coordinate regression} Coordinate regression methods \cite{sun2013deep, zhou2013extensive, toshev2014deeppose, zhang2014coarse, trigeorgis2016mnemonic, lv2017deep, valle2019face} directly predict 2D coordinates of facial landmarks, usually adopting coarse-to-fine approaches with cascaded regression trees or networks. \cite{feng2018wing} introduces wing loss which is less sensitive to outliers. In addition to landmark locations, \cite{chen2019face,gundavarapu2019structured,kumar2020luvli} also measure prediction uncertainties besides their coordinates.

\paragraph{Heatmap regression} Heatmap regression methods \cite{wei2016convolutional,dong2018style,tang2018quantized,deng2019joint,tang2019towards} predict a heatmap for each facial landmark. stacked hourglass (HG) network \cite{newell2016stacked} is the most widely used architecture among recent works \cite{yang2017stacked,bulat2017far,wu2018look}. Apart from the common L$_1$ or L$_2$ loss functions, \cite{wang2019adaptive} adapts wing loss from coordinate regression, and add different foreground and background weights to address the class imbalance in heatmaps. \cite{huang2020propagationnet} introduces focal wing loss to adjust data sample weights. Currently, heatmap regression methods achieve higher accuracy compared to coordinate regression methods in general.

\paragraph{Facial boundary heatmaps } To address the issue of ill-defined contour landmarks, \cite{wu2018look} introduces facial boundary heatmaps which is the heatmaps representation of contours interpolated from facial landmarks. Facial boundary heatmaps are also used in \cite{wang2019adaptive,huang2020propagationnet} to improve landmarks prediction accuracy. In existing works, facial boundary heatmaps have been used only as an internal intermediate representation to mitigate prediction errors caused by ill-defined landmarks. AC differs from facial boundary heatmaps in two aspects: (1) AC aims at modeling true facial contours instead of the interpolated contours from landmarks; and (2) AC can be explicitly extracted as final outputs and can be quantitative evaluated.

\paragraph{Face parsing} Instead of predicting facial landmarks, face parsing methods \cite{liu2015multi,zhou2015interlinked,zhou2017face,liu2017face,lin2019face} aim at segmenting faces into semantic face parts. Compared with facial landmarks models, face parsing models provide extra information such as skin masks and face part areas, but as segmentation models, they focus more on regions rather than landmarks or contours.


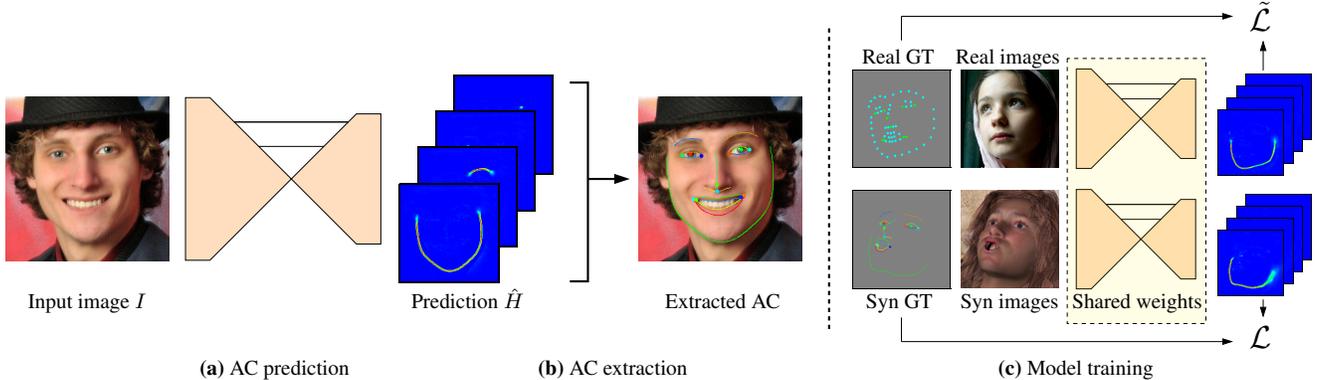
\begin{figure*}[th!]
    \includegraphics[width=\textwidth]{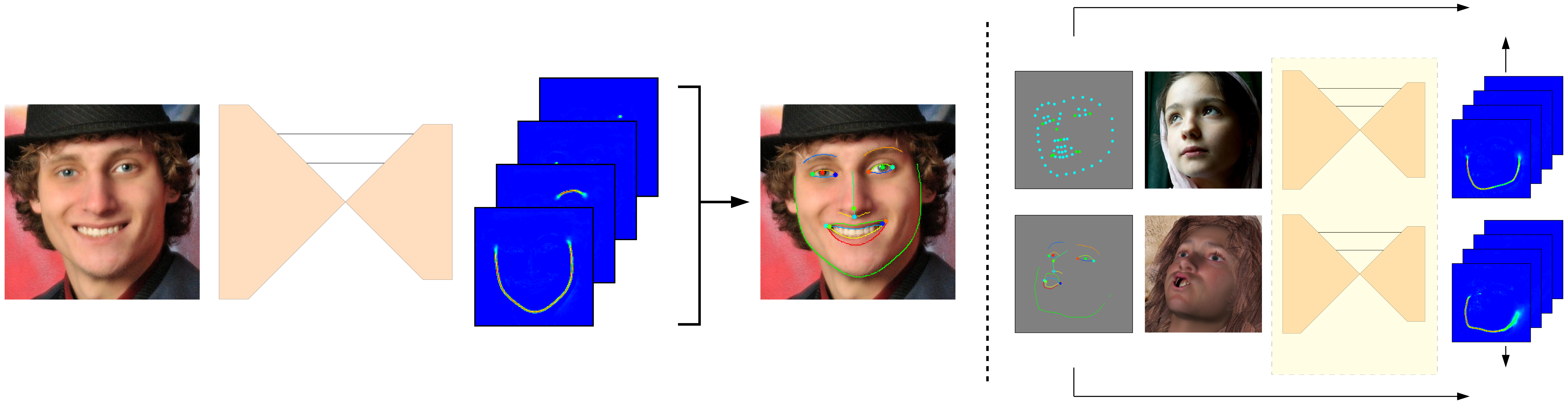}
    \put(-487,15){\footnotesize Input image $I$}
    \put(-342,15){\footnotesize Prediction $\hat{H}$}
    \put(-246,15){\footnotesize Extracted AC}
    \put(-171.5,107.5){\footnotesize Real GT}
    \put(-136,107.5){\footnotesize Real images}
    \put(-170,15){\footnotesize Syn GT}
    \put(-134.3,15){\footnotesize Syn images}
    \put(-92,15){\footnotesize Shared weights}
    \put(-25,120){\large $\Tilde{\mathcal{L}}$}
    \put(-25,0){\large $\mathcal{L}$}
    \put(-422,-10){\footnotesize \textbf{(a)} AC prediction}
    \put(-294,-10){\footnotesize \textbf{(b)} AC extraction}
    \put(-120,-10){\footnotesize \textbf{(c)} Model training}
    \caption{\textbf{Overview of our ACE-Net framework}. (a) AC prediction module that estimates the location of each anchor and contour with a heatmap. (b) AC extraction module that converts heatmaps into anchor coordinates and contours. (c) ACE-Net training framework. Fully supervised losses $\mathcal{L}$ are applied to synthetic data where full AC annotations are available, while weakly supervised losses $\Tilde{\mathcal{L}}$ are applied to real data where only landmarks annotations are available. Best viewed in color.}
    \label{fig:overview}
\end{figure*}

\section{Methods}

\figref{fig:overview} shows an overview of our proposed ACE-Net framework. At test time (\figref{fig:overview}a,b), ACE-Net takes a face ROI image as the input and first predicts a separate heatmap for each facial anchor and contour, and then performs AC extraction to locate each anchor and contour with sub-pixel accuracy.
At training time (\figref{fig:overview}c), ACE-Net is trained with both synthetic and real images. The synthetic images are rendered from 3D face meshes and thus the Ground Truth (GT) ACs are automatically available, while the real images only have annotated GT landmarks. We apply different losses to the synthetic and real data based on the different information available from each. For real data, we introduce a weakly supervised loss function which encourages the predicted contours to pass through available annotated GT landmarks. For the synthetic data, we use the fully supervised loss since complete GT ACs are available.

In this section we first define our AC representation followed by the definitions of our fully supervised and weakly supervised loss functions used to train the AC prediction module. Finally, we describe our AC extraction module that enables explicit AC localization.

\subsection{The Anchor-Contour (AC) Representation}

\paragraph{Anchor} An anchor, $a$, is point feature tied to well defined local features on the face, e.g., corners of the eyes or lips. It is represented as a 2D point, $(a_x, a_y)$, on the image. 

From GT anchor landmarks, anchor heatmaps are generated for training our network. The heatmap, $H_a$, corresponding to an anchor $a$ is defined as
\begin{equation}
    H_a(p) = \max\left(0,\; 1 - 2 \; \frac{\norm{p - a}^2}{\sigma^2}\right)
    \label{eq:def_anchor_heat}
\end{equation}
where $p$ is a pixel on the heatmap and $\sigma$ controls the width of the peak. 
We choose this definition of anchor heatmaps instead of the commonly used Gaussian heatmaps to keep it consistent with our contour heatmaps definition (below). 

\paragraph{Contour} A facial contour, $c$, is a 2-d curve that maps to well-defined facial contours, e.g., eyelids, lips. The occluding contour on the face (outer boundary) is also represented as a facial contour despite the fact that these are related to the viewing direction and not tied to specific features on the face. The contour, $c$, is represented as a collection of sub-pixel line segments, $\{s_i\}$. 

Similar to anchors, heatmaps are generated from contours for training our network. A contour heatmap $H_c$ is define as
\begin{equation}
    H_c(p) = \max\left(0,\; 1 - 2 \; \frac{dist(p, c)^2}{\sigma^2}\right)
    \label{eq:def_contour_heat}
\end{equation}
where $p$ is any pixel on the heatmap, and $dist(p,c)$ is a function measuring the minimum distance from point $p$ to the contour $c$, specifically, the minimum distance between $p$ and any of the line segments $s_i$ in $c$. 

\subsection{Fully Supervised Loss}
In the fully supervised setting, for each image $I$, its anchors $A = \{a_1, a_2, \dots, a_{N_A}\}$ and contours $C = \{c_1, c_2, \dots, c_{N_C}\}$ are all available. We generate the ground truth AC heatmaps $H =\{H_a^1, \dots, H_a^{N_A}, H_c^1, \dots, H_c^{N_C}\}$ following \eqref{eq:def_anchor_heat} and \eqref{eq:def_contour_heat}, and define the fully supervised loss $\mathcal{L}$ for predicted heatmaps $\hat{H}$ as a weighted root mean square loss:
\begin{equation}
    \mathcal{L}(H, \hat{H}) = \dfrac{1}{|H|} \sqrt{\sum W(H, \hat{H}) \cdot \norm{H - \hat{H}}^2}
    \label{eq:loss_full_heat}
\end{equation}
where $|H|$ is the total number of pixels in heatmaps H, $\cdot$ is element-wise multiplication, and $W$ is a weighting emphasizing positive and hard negative examples:
\begin{equation}
    W(H, \hat{H}) = 1 + (\alpha - 1) \cdot \max \left(H,\; \norm{H - \hat{H}}\right)
\end{equation}
This weighting is necessary because most of the pixels in heatmaps are background pixels. Without $W_h$ the learned model is easily trapped in the trivial local minimum $\hat{H} = 0$.

In settings where line-contours, $\Tilde{C}$, are generated by connecting GT contour landmarks with straight line segments, since $\Tilde{C} \ne C$, applying this fully supervised loss $\mathcal{L}$ with $\Tilde{C}$ being the ground truth will result in significant loss at the fine-grained level.

\subsection{Weakly Supervised Loss}
In the weakly supervised setting, for each image, $I$, only its landmarks, $L = \{l_1, l_2, \dots, l_{N_L}\}$, are available. For predicted anchor heatmaps, $\hat{H}_a$, we just apply the fully supervised loss, $\mathcal{L}$, in \eqref{eq:loss_full_heat}, because anchors are a subset of landmarks and we have access to all ground truth anchors $A$. However, the weakly supervised contour loss, $\Tilde{\mathcal{L}}$, needs a novel treatment. 

\begin{figure}[!t]
    \captionsetup[subfigure]{belowskip=-5pt}
    \begin{subfigure}[b]{0.10\textwidth}
        \centering
        \includegraphics[height=40pt]{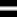}
        \caption{$\theta = 0\degree$}
    \end{subfigure}
    \hfill
    \begin{subfigure}[b]{0.10\textwidth}
        \centering
        \includegraphics[height=40pt]{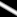}
        \caption{$\theta = 30\degree$}
    \end{subfigure}
    \hfill
    \begin{subfigure}[b]{0.10\textwidth}
        \centering
        \includegraphics[height=40pt]{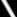}
        \caption{$\theta = 60\degree$}
    \end{subfigure}
    \hfill
    \begin{subfigure}[b]{0.10\textwidth}
        \centering
        \includegraphics[height=40pt]{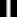}
        \caption{$\theta = 90\degree$}
    \end{subfigure}
    \caption{Examples of contour template $T_{\sigma,\theta}$ with various $\theta$.}
    \label{fig:contour_template}
\end{figure}

Before we define our weakly supervised contour loss, $\Tilde{\mathcal{L}}$, we first introduce a \textbf{``contourness score''}, $\mathcal{C}$, that evaluates whether there is a contour passing through a given pixel $p$ on a heatmap $H$. Assuming that a contour is locally a straight line with orientation $\theta$ and width $\sigma$, we construct its heatmap template $T_{\sigma,\theta}$ based on \eqref{eq:def_contour_heat}:
\begin{equation}
    T_{\sigma,\theta}(x,y) = \max\left(0, 1 - 2 \; \frac{(y\cos\theta - x\sin\theta)^2}{\sigma^2}\right)
    \label{eq:def_contourness_template}
\end{equation}
Examples of $T_{\sigma,\theta}$ is shown in \figref{fig:contour_template}. If this contour passes through pixel $p$, then heatmap $H$ cropped around $p$ should match the template $T_{\sigma,\theta}$. Thus, based on template matching error, we define contourness score, $\mathcal{C}$,  as
\begin{align}
    \begin{split}
        \mathcal{C}_{\sigma}(H,p) = &-\min_{\theta} \sum_{i=-2\sigma}^{2\sigma} \sum_{j=-2\sigma}^{2\sigma} G_{\sigma}(i,j) \cdot \\
        &\norm{H_+(p_x+i, p_y+j) - T_{\sigma,\theta}(i, j)}^2
    \end{split}
    \label{eq:def_contourness_pixel}
\end{align}
Here the negative sign at the beginning turns contourness $\mathcal{C}$ from an error to a score. $H_+ = \max(0,H)$ is a non-negative clipping, and $G_{\sigma}(x,y) = e^{-(x^2+y^2) / \sigma^2}$ is the Gaussian weight. 
We will omit $\sigma$ in the subscripts since it is a constant hyper-parameter.

The optimization in \eqref{eq:def_contourness_pixel} has a closed-form solution through the use of steerable filters \cite{freeman1991design}. Therefore, the contourness score, $\mathcal{C}(H)$, for all pixels on heatmap $H$ can be efficiently computed with convolutions and integrated into loss functions. Due to space limitation, please refer to supplementary materials for detailed description of this process. Apart from the contourness score map $\mathcal{C}(H)$, contour orientation map, $\mathcal{O}(H)$, can also be efficiently computed as the optimal $\theta$ in the closed-form solution to \eqref{eq:def_contourness_pixel}.

Now, with the contourness score $\mathcal{C}$ defined, we can begin to define the weakly supervised loss. Recall that in this setting we only have the GT landmarks to work with. Let $\Tilde{c}$ be the a line-contour created from corresponding contour landmarks, $L_c$. Note that $\Tilde{c}$ can either be generated by simply connecting the adjacent GT landmarks with straight lines, or it can be generated by interpolating the landmarks with splines. Let $\hat{H}$ be a predicted contour heatmap representing a predicted contour, $\hat{c}$. 

We design our weakly supervised loss to enforce the following three rules:

(1) The predicted contour $\hat{c}$ must pass through all contour landmarks $L_c$:
\begin{equation}
    \Tilde{\mathcal{L}}_{landmark}(\hat{H}) = \dfrac{1}{|L_c|} \sum_{l \in L_c} f\left( \mathcal{C}(\hat{H}, l) \right)
    \label{eq:def_loss_weak_landmark}
\end{equation}
where $|L_c|$ is the total number of contour landmarks, and $f(\cdot)$ is a mapping function that converts contourness score to a loss in $[0,1]$ range.

(2) The predicted contour $\hat{c}$ must be close to line-contour $\Tilde{c}$. In other words, for each pixel $p$ on $\Tilde{c}$, there must exist a pixel $q$ on $\hat{c}$ such that $q-p$ is the line-contour normal at $p$, and $\norm{p-q} \le D$ for some constant threshold $D$:
\begin{equation}
    \Tilde{\mathcal{L}}_{line}(\hat{H}) = \dfrac{1}{|\Tilde{c}|} \sum_{p \text{ on } \Tilde{c}} f\left( \max_{-D \le d \le D} \mathcal{C}(H, \; p + d \cdot \mathcal{N}_{\Tilde{c}}(p)) \right)
    \label{eq:def_loss_weak_line}
\end{equation}
where $|\Tilde{c}|$ is the total number of pixels on $\Tilde{c}$, $\mathcal{N}_{\Tilde{c}}$ is the normal map of line-contour $\Tilde{c}$ and $\mathcal{N}_{\Tilde{c}}(p)$ is the line-contour normal at pixel $p$.

(3) Pixels far away from line-contour $\Tilde{c}$ should have zero heat value:
\begin{equation}
    \Tilde{\mathcal{L}}_{far}(\hat{H}) = \dfrac{1}{|H|} \sqrt{\sum M(\Tilde{c}) \cdot W(0, \hat{H}) \cdot \norm{\hat{H}}^2}
    \label{eq:def_loss_weak_far}
\end{equation}
where $M(\Tilde{c})$ is a binary mask selecting pixels far from $\Tilde{c}$:
\begin{equation}
    M(\Tilde{c},p) = 
    \begin{cases}
      1, & dist(p,\Tilde{c}) > D \\
      0, & \text{otherwise}
    \end{cases}
    \label{eq:def_mask_far}
\end{equation}
\eqref{eq:def_loss_weak_far} follows $\mathcal{L}$ in \eqref{eq:loss_full_heat} with ground truth being $H = 0$. It only adds an extra mask $M(\Tilde{c})$ to select pixels far from $\Tilde{c}$.

Finally, we define our weakly supervised contour loss $\Tilde{\mathcal{L}}$ as the sum of the three losses above:
\begin{equation}
    \Tilde{\mathcal{L}} = \Tilde{\mathcal{L}}_{far} + \lambda_{landmark} \cdot \Tilde{\mathcal{L}}_{landmark} + \lambda_{line} \cdot \Tilde{\mathcal{L}}_{line}
    \label{eq:loss_weak_contour}
\end{equation}
where $\lambda_{landmark}$ and $\lambda_{line}$ are constant weights.

\subsection{AC Extraction}
The AC Extraction module converts each predicted anchor heatmap, $H_a$, to a 2D anchor position $a = (a_x, a_y)$, and each contour heatmap, $H_c$, into to a contour, $c = \{s_i\}$.

\paragraph{Extracting Anchors}
We adopt the local center-of-mass method: given an anchor heatmap $H_a$, we find the pixel $p^*$ with highest heat value, and compute anchor position $a$ as
\begin{equation}
    a = \sum\limits_{p:\; \norm{p-p^*} \le \sigma} H_a(p) \cdot p \bigg/ \sum\limits_{p:\; \norm{p-p^*} \le \sigma} H_a(p)
\end{equation}

\begin{figure}[!tp]
    \captionsetup[subfigure]{belowskip=-5pt}
    \begin{subfigure}[b]{0.17\textwidth}
        \includegraphics[width=\textwidth]{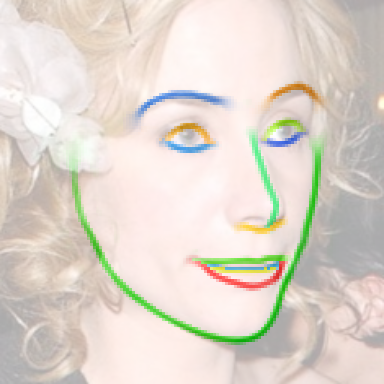}
        \caption{Heatmaps $H$}
    \end{subfigure}
    \hfill
    \begin{subfigure}[b]{0.1\textwidth}
        \includegraphics[trim=0 28 0 32, clip, width=\textwidth]{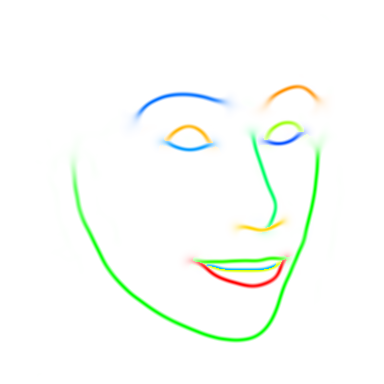}
        \includegraphics[trim=0 28 0 28, clip, width=\textwidth]{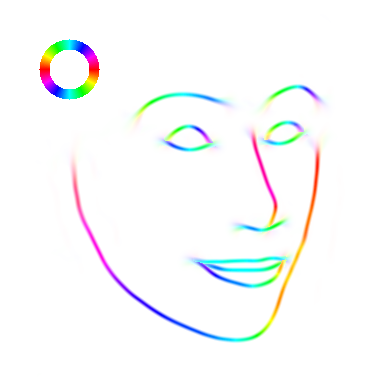}
        \caption{$\mathcal{C}$ and $\mathcal{N}$}
    \end{subfigure}
    \hfill
    \begin{subfigure}[b]{0.17\textwidth}
        \includegraphics[width=\textwidth]{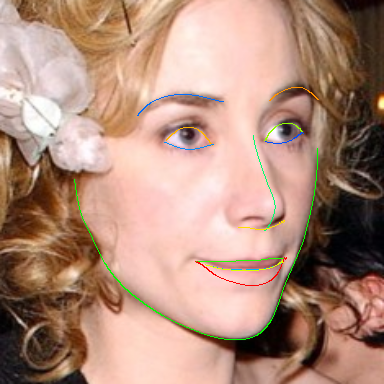}
        \caption{Extracted contours}
    \end{subfigure}
    \caption{Contour extraction. (a) Contour heatmaps predicted by ACE-Net visualized on top of the input image. Due to space limitation we visualize heatmaps for all contours in one image with hue indicating contour ID and saturation indicating heat values. (b) Top: contourness map $\mathcal{C}$; bottom: color coded normal map $\mathcal{N}$ with normal orientation being hue and contourness being saturation. (c) Extracted contour overlaid on the input image. Color indicates contour ID.}
    \label{fig:AC_extraction}
\end{figure}

\paragraph{Extracting Contours} Given a contour heatmap, $H_c$, we compute its contourness map, $\mathcal{C}(H_c)$, as well as its contour orientation map, $\mathcal{O}(H_c)$, using the closed-form solution to \eqref{eq:def_contourness_pixel}. We then obtain the contour normal map $\mathcal{N}(H_c) = \mathcal{O}(H_c) + \pi / 2$, and perform non-maximum suppression (NMS) on $\mathcal{C}(H_c)$ along directions specified by $\mathcal{N}(H_c)$, retaining just the maximal pixels which are then thresholded to obtain the binary contour mask, $B_c$. During NMS, we also localize the points in $B_c$ to subpixel accuracy by fitting a parabola along the normal direction specified by $\mathcal{N}(H_c)$. After NMS and thresholding, each contour has a width of 1 pixel. Connected components analysis is performed with hysteresis to extract the contour trace, $c$, in the same way as Canny edge detection. (\figref{fig:AC_extraction}).  

\section{Synthetic Data Generation}
Our synthetic training data is generated using a 3D morphable face model (3DMM) \cite{blanz1999morphable} created by \cite{facegen}. We randomly select facial shape, expression and texture coefficients for the 3DMM and generate 50,000 3D face meshes with random accessories (hairs, glasses and headbands). Each mesh is rendered to a image with random lighting and head pose up to $60\degree$ yaw, $30\degree$ pitch and $30\degree$ roll. An image from non-human categories of Caltech-256 dataset \cite{griffin2007caltech} is randomly picked for each synthetic face at training time as the background.

\begin{figure}[!tbp]
    \includegraphics[width=0.11\textwidth]{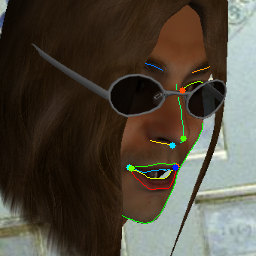} \hfill
    \includegraphics[width=0.11\textwidth]{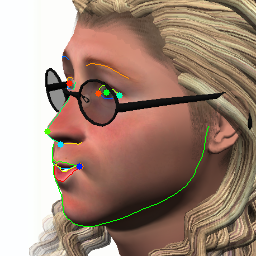} \hfill
    \includegraphics[width=0.11\textwidth]{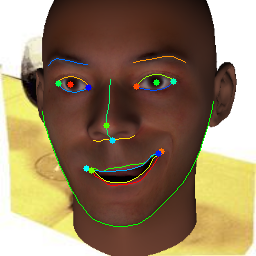} \hfill
    \includegraphics[width=0.11\textwidth]{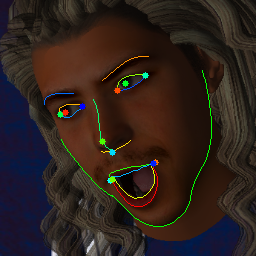} \\
    \vspace{-3mm}\\
    \includegraphics[width=0.11\textwidth]{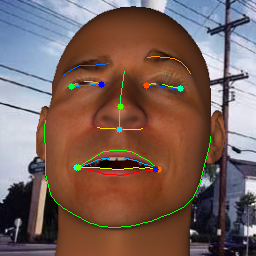} \hfill
    \includegraphics[width=0.11\textwidth]{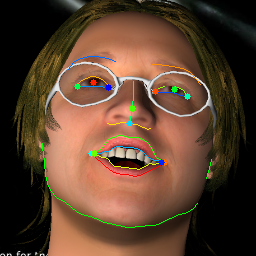} \hfill
    \includegraphics[width=0.11\textwidth]{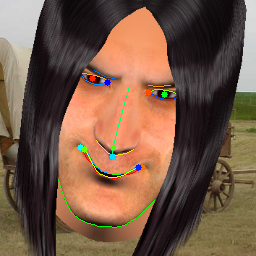} \hfill
    \includegraphics[width=0.11\textwidth]{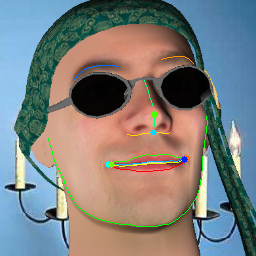}
    \caption{Generated synthetic images and their AC annotations.}
    \label{fig:synthetic_data}
\end{figure}

To obtain AC annotations for the synthetic data, we manually annotate the indices of all mesh vertices in the 3DMM corresponding to each facial anchor and contour, and then automatically generate the AC annotation for each synthetic image using the generated morph coefficients and head poses.  Example synthetic images and their annotations are shown in \figref{fig:synthetic_data}.


\section{Experiments}
\begin{figure}[!b]
    \begin{subfigure}[b]{0.15\textwidth}
        \includegraphics[trim=32 32 32 32, clip, width=\textwidth]{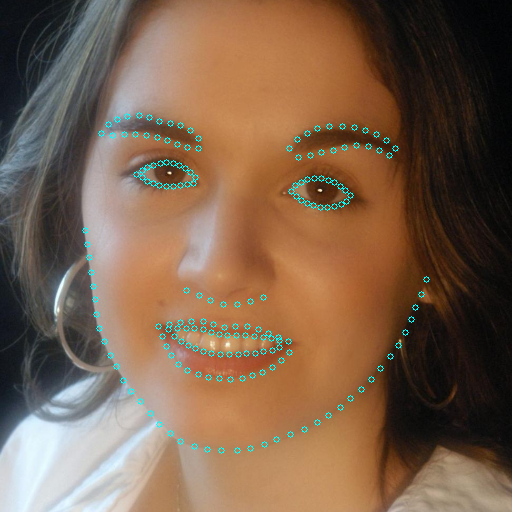}
        \caption{HELEN (194)}
        \label{fig:annotations_HELEN}
    \end{subfigure}
    \hfill
    \begin{subfigure}[b]{0.15\textwidth}
        \includegraphics[trim=32 32 32 32, clip, width=\textwidth]{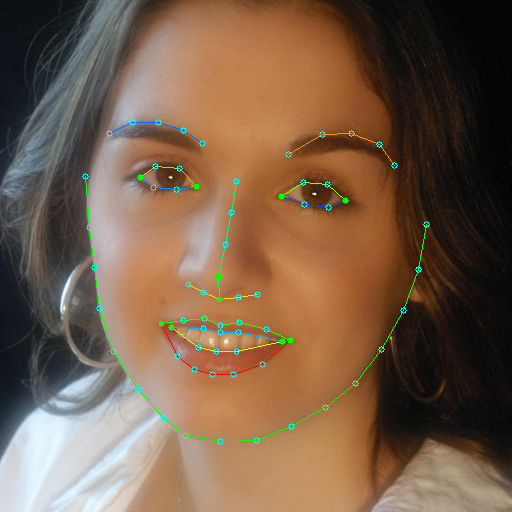}
        \caption{300-W (68)}
        \label{fig:annotations_300W}
    \end{subfigure}
    \hfill
    \begin{subfigure}[b]{0.15\textwidth}
        \includegraphics[trim=32 32 32 32, clip, width=\textwidth]{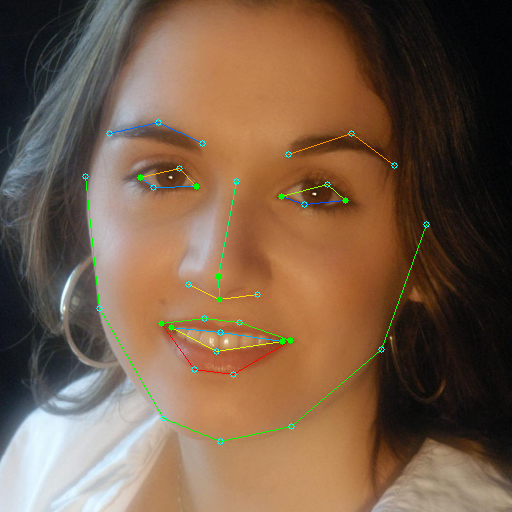}
        \caption{sparse (36)}
        \label{fig:landmarks_annotations}
    \end{subfigure}
    \caption{Comparison of landmarks annotations. (a) 194-landmark annotation in \textbf{HELEN} used for evaluation. (b) 68-landmark annotation in \textbf{300-W} used to train AC based models. (c) 36-landmark sparse annotation used for ablation study.}
    \label{fig:dataset_landmarks}
\end{figure}

Following the assumption that accurate contour annotations are hard to obtain, we train our models with the most widely used facial landmarks dataset, 300-W, and evaluate the fine-level face alignment accuracy on the HELEN dataset where extremely dense landmarks annotations are available to approximate true facial contours. \figref{fig:dataset_landmarks} illustrates the landmarks annotation density of the two datasets.

\subsection{Datasets}
\paragraph{300-W}\cite{sagonas2013300} is by far the most widely used face alignment dataset where each image is annotated with 68 facial landmarks. Its training set consists of 3148 images.

\paragraph{HELEN}\cite{le2012interactive} testset contains 330 test images, each annotated with 194 facial landmarks. We choose HELEN as our evaluation data because it is the only publicly available dataset with dense enough landmarks annotation to approximate true facial contours. We do not include HELEN training set for evaluation, because it overlaps with our training dataset 300-W.

\subsection{Evaluation Metrics}
\paragraph{Normalized Mean Error (NME\textsuperscript{AC})} Following the standard NME for facial landmarks, we define a similar metric, NME\textsuperscript{AC}, for anchors and contours. Note that our ground truths are just landmarks, albeit a denser set than what was used to train ACE-Net. We first divide these ground truth landmarks, $L$, into anchor landmarks, $L^A = \{a_1, a_2, \dots\}$, and contour landmarks, $L^C = \{L_1^C, L_2^C, \dots\}$ such that $L_i^C \ $ is the set of landmarks on ground truth contour $c_i$. Then, given the predicted anchors, $\hat{A} = \{\hat{a}_1, \hat{a}_2, \dots\}$, and contours, $\hat{C} = \{\hat{c}_1, \hat{c}_2, \dots\}$, we define the prediction error of each GT landmark, $l$, as:
\begin{equation}
    err(l,\hat{A},\hat{C}) = \begin{cases}
        \norm{l - \hat{a}_i}_2, & \text{if } \exists i, l = a_i \in L^A\\
        dist(l, \hat{c}_i), & \text{if } \exists i, l \in L_i^C.
    \end{cases}
\end{equation}
where $dist(\cdot,\cdot)$ is the same point-to-contour distance as in \eqref{eq:def_contour_heat}.

Finally, the metric NME\textsuperscript{AC} is defined as the normalized mean error of all landmarks:
\begin{equation}
    \text{NME\textsuperscript{AC}}(L,\hat{A},\hat{C}) = \frac{100\%}{d} \cdot \frac{1}{N_L} \sum\limits_{i=1}^{N_L} err(l_i,\hat{A},\hat{C})
\end{equation}
where $d$ is the normalization factor. For all experiments in this paper, we use inter-ocular distance (distance between the two outer eye corners) as the normalization factor.

\paragraph{Area Under the Curve (AUC)} measures the area under the cumulative error distribution (CED) curve with the error being NME\textsuperscript{AC}. We cut off the error at $6\%$.

\subsection{Implementation Details}
We obtain face bounding boxes by padding the tight boxes from ground truth landmarks by 25\%, 25\%, 33\%, 17\% on left, right, top and bottom sides respectively, and then convert them into square boxes by keeping the longest edge length. Our training data augmentation includes random translation up to $25\%$, rotation up to $30\degree$, horizontal flipping with $50\%$ chance, gamma correction with $\gamma \in (2^{-0.75}, 2^{0.75})$ and Gaussian noise. We set anchor and contour width $\sigma=2$, and contourness pooling radius $D=6$ for our weakly supervised losses in \eqref{eq:def_loss_weak_line} and \eqref{eq:def_mask_far}. 
We choose the mapping function $f(\mathcal{C}) = 1 - 2^{(\mathcal{C} - 4.92)/1.5}$ for \eqref{eq:def_loss_weak_landmark} and \eqref{eq:def_loss_weak_line} since the maximum contourness $\mathcal{C}$ is approximately 4.92 when $\sigma=2$. We empirically set weights $\lambda_{landmark} = \lambda_{line} = 0.1$.

The CNN used in our experiments has an input size of $256 \times 256$ (grayscale) and an output size of $128 \times 128 \times (N_A + N_C)$ where $N_A$ and $N_C$ are the number of anchors and contours, respectively. Details of our network architecture are described in our supplementary materials due to space limitations. ACE-Net does not rely on specific network architecture, so it can be replaced by any other commonly used CNNs.
We train our models from scratch using Adam optimizer \cite{kingma2014adam} with an initial learning rate of $2.5 \times 10^{-4}$, fuzz factor at $10^{-8}$ and learning rate decay at $10^{-6}$. All our models are trained with three 8GB GeForce GTX 1080 GPUs.

At test time, we use $\sigma=3$ for AC extraction. We choose a larger $\sigma$ than in training because the predicted heatmaps tend to be more ``spread out" than the ground truth heatmaps.

\subsection{Evaluation on the HELEN Dataset}
In this paper, all models are trained on 300-W training set with 68 landmarks per image, and the resulting AC representations are evaluated on HELEN test set with 194 landmarks per image. During evaluation, eyebrow boundary landmarks and nose side landmarks are excluded since they are not annotated in 300-W. We compare the performance of the proposed ACE-Net with the following models:

\paragraph{Lmk-line} first predicts 68 facial landmarks through conventional heatmap regression, and then converts the predicted landmarks into anchors and line-contours. Lmk-line has the same backbone architecture as ACE-Net, and its training loss is $\mathcal{L}$ in \eqref{eq:loss_full_heat}. Note that this is just for baseline reference and is not at SOTA-level.

\paragraph{GT68-line} converts the ground truth 68 facial landmarks (from the 300-W testset which are also available in the HELEN testset) into anchors and line-contours. We use the ground truth landmarks instead of any particular SOTA landmark extraction results because the GT landmarks represent the upper bound for landmarks detection accuracy and consequently these line-contours must represent an upper bound for landmarks-based line-contour models.

\paragraph{GT68-interp} is similar to GT68-line but uses quadratic spline interpolation instead of straight lines to generate contours from landmarks. We also tried cubic interpolation but it is more sensitive to landmark noise than quadratic interpolation and so does not perform as well. 

\paragraph{Line} predicts AC with the same architecture as ACE-Net, but is trained only with real images and treats line-contours obtained from GT landmarks as GT contours.

\paragraph{Line$^*$} extends Line by including synthetic data in its training set.

To sum up, \textbf{Lmk-line}, \textbf{GT68-line} and \textbf{GT68-interp} are landmarks-based models with no special training to extract contours, while \textbf{Line}, \textbf{Line$^*$} and \textbf{ACE-Net} are AC based models, specifically trained to extract contours. The differences between these AC based models are only in the training data and losses:
\begin{center}
    \begin{tabular}{ |c|c|c| } 
        \hline
         Model      & Training data & Loss \\ 
        \hline
         Line       & Real  & $\mathcal{L}$  \\ 
         Line$^*$   & Real + Syn & $\mathcal{L}$ on Syn + ${\mathcal{L}}$ on Real \\ 
         ACE-Net    & Real + Syn & $\mathcal{L}$ on Syn + $\hat{\mathcal{L}}$ on Real \\ 
        \hline
    \end{tabular}
\end{center}
Here ``Real" stands for the 300-W training set and ``Syn" stands for the generated synthetic data.

\begin{table}[!tp]
    \centering
    \begin{tabular}{ |c|c|c|c|c|c| } 
        \hline
        Model     & Overall & Eyes & Nose & Mouth & Chin \\ 
        \hline
        Lmk-line    & 2.20 & 1.52 & 2.06 & 2.79 & 2.09 \\ 
        GT68-line   & 1.48 & 1.05 & 1.84 & 1.37 & 1.99 \\ 
        GT68-interp & 1.39 & 0.96 & 1.84 & \textbf{1.26} & 1.91 \\ 
        Line        & 1.59 & 1.23 & 1.15 & 1.35 & 2.34 \\ 
        Line$^*$    & 1.59 & 1.17 & \textbf{1.14} & 1.32 & 2.45 \\ 
        ACE-Net     & \textbf{1.32} & \textbf{0.94} & 1.17 & 1.29 & \textbf{1.77} \\ 
        \hline
    \end{tabular}
    \caption{NME\textsuperscript{AC} on HELEN testset}
    \label{table:NME_HELEN}
\end{table}

\begin{figure}[!tp]
    \centering
    \includegraphics[trim=20 0 40 20, clip, width=0.47\textwidth]{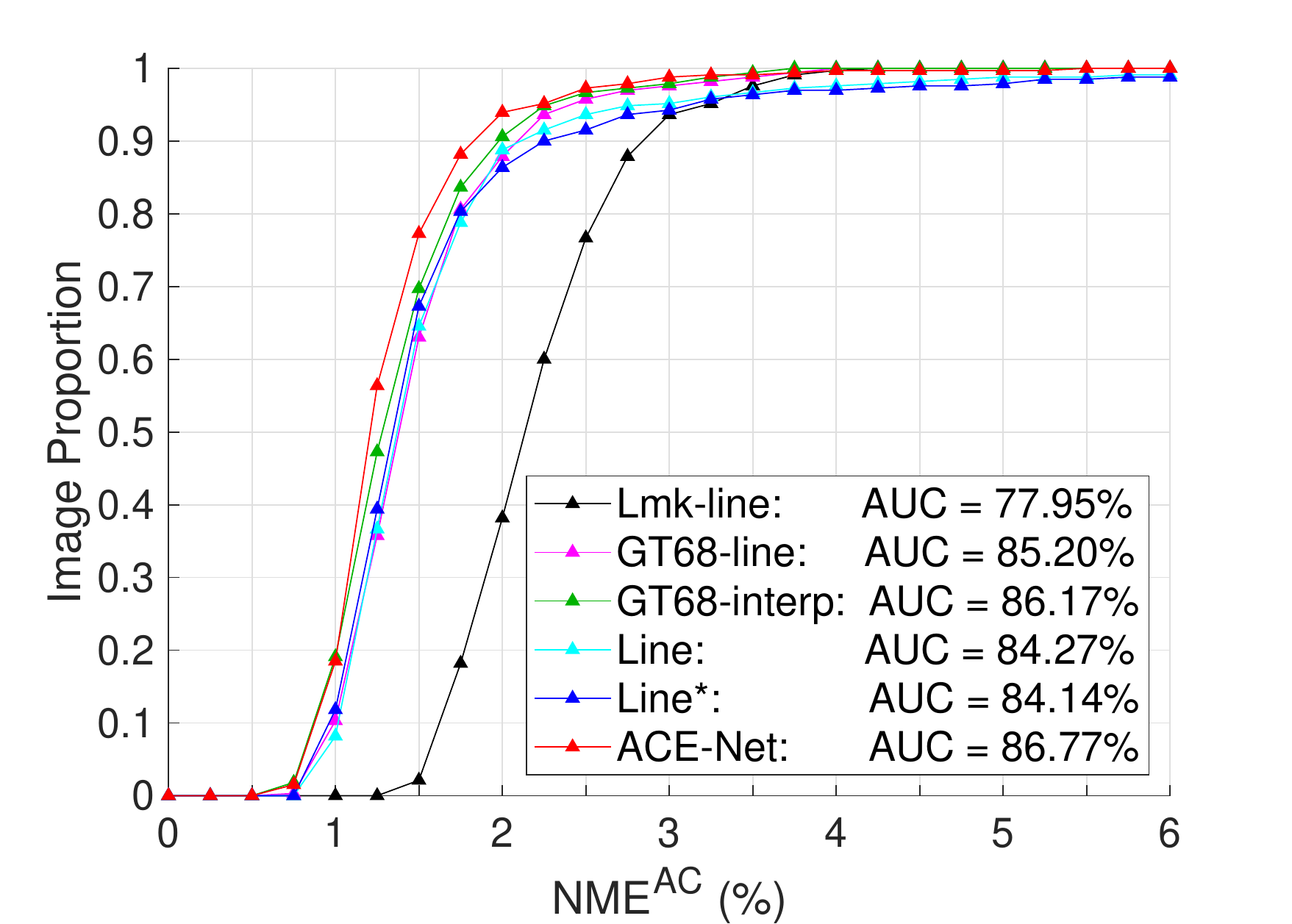}
    \caption{CED curve on HELEN testset}
    \label{fig:CED_HELEN}
\end{figure}

The results on HELEN testset are shown in \tabref{table:NME_HELEN} and \figref{fig:CED_HELEN}. \textbf{ACE-Net} achieves significantly better accuracy than its landmarks detection based counterpart \textbf{Lmk-line} in all face parts (NME\textsuperscript{AC}: $1.32$ vs $2.20$, AUC: $86.77\%$ vs $77.95\%$). Notably, \textbf{ACE-Net} also outperforms ground truth landmarks based models \textbf{GT68-line} and \textbf{GT68-interp} (NME\textsuperscript{AC}: $1.32$ vs $1.48$ and $1.39$, AUC: $86.77\%$ vs $85.20\%$ and $86.17\%$), showing the effectiveness of AC based modeling at fine-level face alignment.

Among the AC based models, \textbf{ACE-Net} achieves higher accuracy than \textbf{Line} (NME\textsuperscript{AC}: $1.32$ vs $1.59$, AUC: $86.77\%$ vs $84.27\%$), especially in the small error region. For example, \textbf{ACE-Net} has a success rate of $56.4\%$ at NME\textsuperscript{AC} $\leq 1.25\%$ whereas \textbf{Line} only has $36.7\%$. This is because ACE-Net's weakly supervised loss $\mathcal{L}$ prevents it from overfitting to the inaccurately hallucinated portions of line-contours, and instead learns from accurate synthetic contours. Without the weakly supervised loss $\mathcal{L}$, synthetic data alone does not bring any noticeable improvement, since there is a large domain gap between the synthetic and the real data, and the model tends to overfit to the inaccurate real line-contours. Therefore, model \textbf{Line$^*$} has almost identical performance to \textbf{Line} in spite of the addition of synthetic data (NME\textsuperscript{AC}: $1.59$ vs $1.59$, AUC: $84.14\%$ vs $84.27\%$).

\subsection{Ablation Study}

\begin{table}[!tp]
    \centering
    \begin{tabular}{ |c|c|c|c|c|c| } 
        \hline
        Model     & Overall & Eyes & Nose & Mouth & Chin \\ 
        \hline
        ACE-Net     & 1.32 & 0.94 & 1.17 & 1.29 & 1.77 \\
        \hline
        GT36-line   & 2.42 & 1.69 & 1.92 & 1.94 & 3.89 \\ 
        GT36-interp & 1.62 & 1.05 & 1.72 & 1.34 & 2.54 \\ 
        ACE-Net\textsuperscript{sparse} & \textbf{1.38} & \textbf{0.96} & \textbf{1.23} & \textbf{1.31} & \textbf{1.92} \\ 
        \hline
    \end{tabular}
    \caption{NME\textsuperscript{AC} on HELEN testset}
    \label{table:NME_HELEN_sparse}
\end{table}

\begin{figure}[!tp]
    \centering
    \includegraphics[trim=20 0 40 20, clip, width=0.47\textwidth]{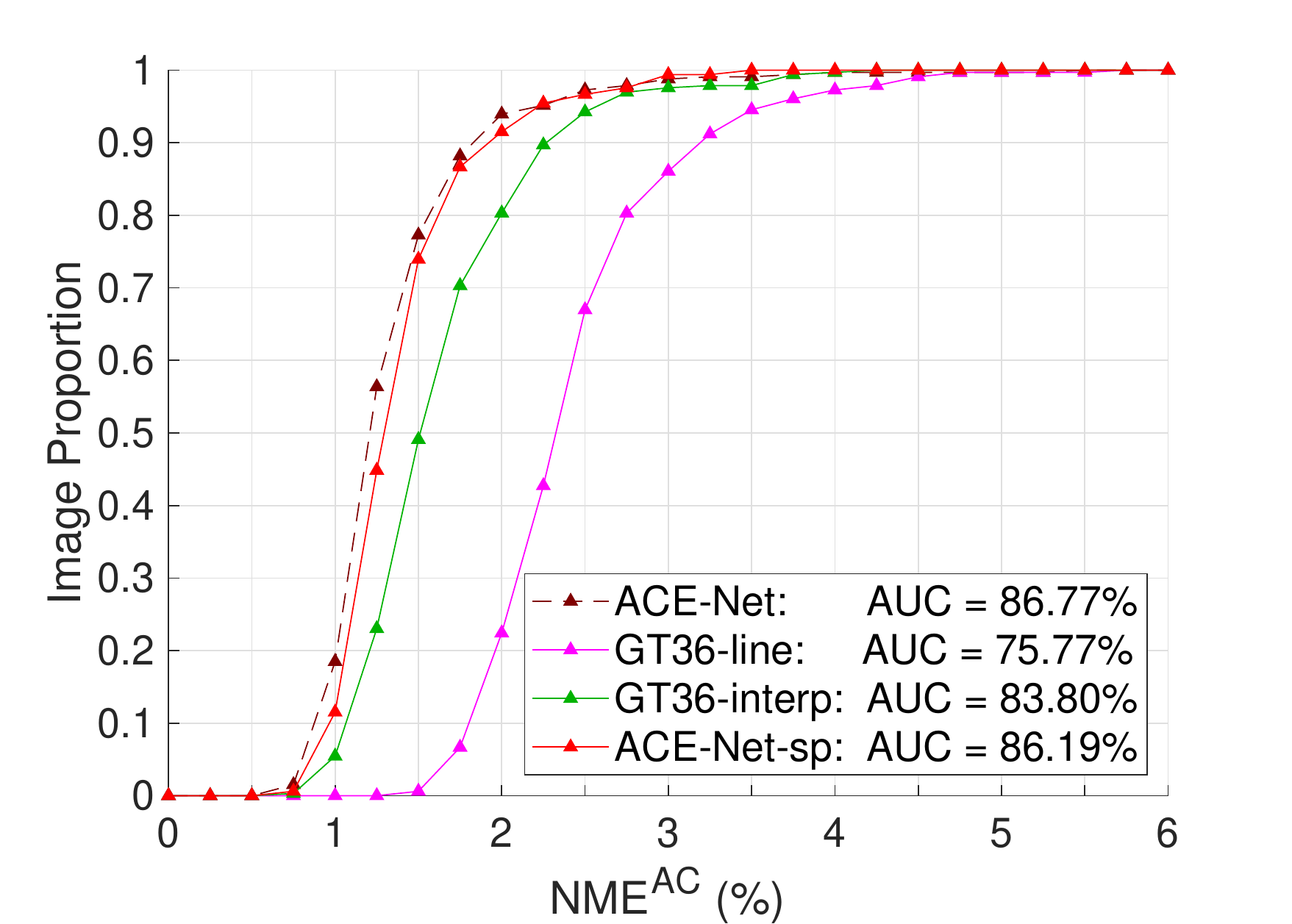}
    \caption{CED curve on HELEN testset. $\text{ACE-Net}^{\text{sparse}}$ is labelled as ``ACE-Net-sp" in the figure due to space limitation.}
    \label{fig:CED_HELEN_sparse}
\end{figure}

\begin{figure*}[!tp]
    \includegraphics[width=0.12\textwidth]{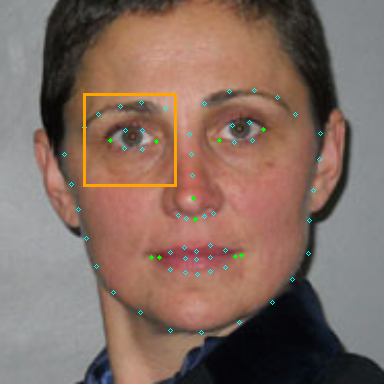} 
    \includegraphics[width=0.12\textwidth]{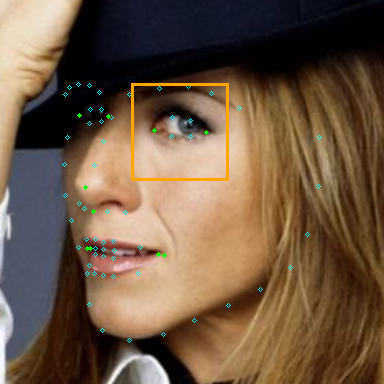} 
    \includegraphics[width=0.12\textwidth]{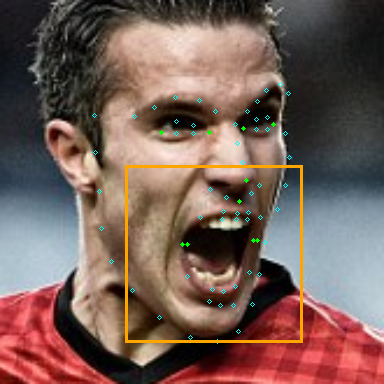} 
    \includegraphics[width=0.12\textwidth]{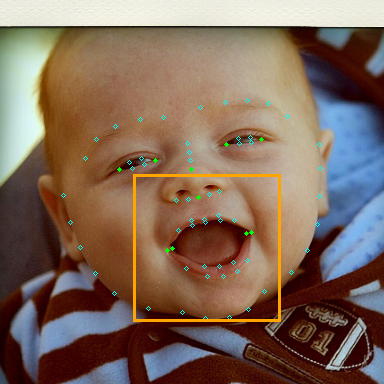} 
    \includegraphics[width=0.12\textwidth]{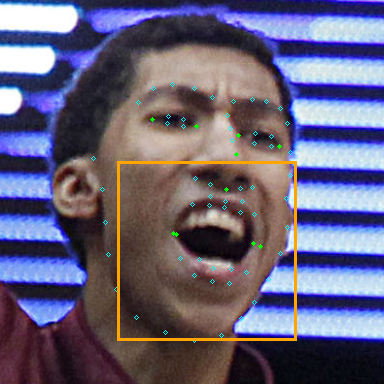} 
    \includegraphics[width=0.12\textwidth]{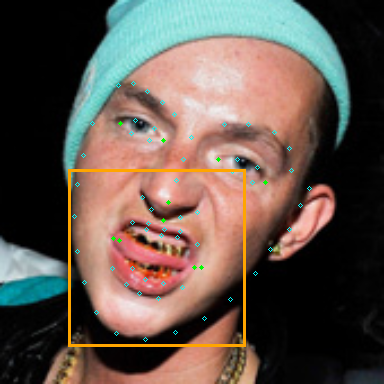} 
    \includegraphics[width=0.12\textwidth]{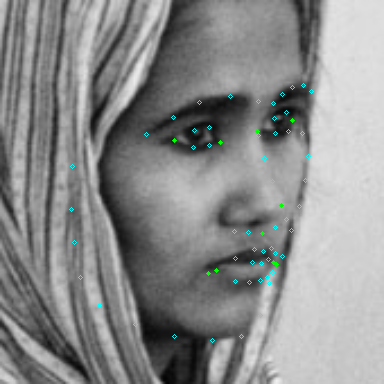} 
    \includegraphics[width=0.12\textwidth]{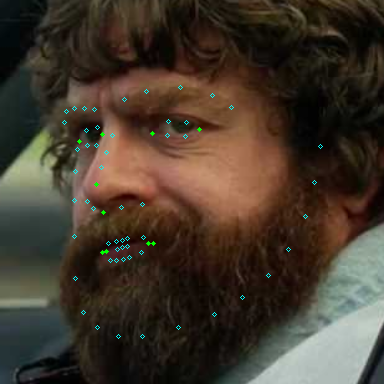} 
    \put(-491,3){\footnotesize \textcolor{white}{GT}} 
    \put(-429,3){\footnotesize \textcolor{white}{GT}} 
    \put(-367,3){\footnotesize \textcolor{white}{GT}} 
    \put(-305,3){\footnotesize \textcolor{white}{GT}} 
    \put(-243,3){\footnotesize \textcolor{white}{GT}} 
    \put(-181,3){\footnotesize \textcolor{white}{GT}} 
    \put(-119,3){\footnotesize \textcolor{white}{GT}} 
    \put( -57,3){\footnotesize \textcolor{white}{GT}} 
    \\
    \includegraphics[width=0.12\textwidth]{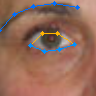} 
    \includegraphics[width=0.12\textwidth]{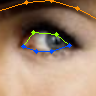} 
    \includegraphics[width=0.12\textwidth]{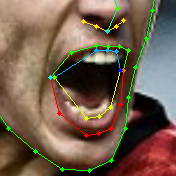} 
    \includegraphics[width=0.12\textwidth]{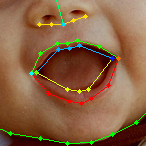} 
    \includegraphics[width=0.12\textwidth]{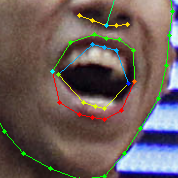} 
    \includegraphics[width=0.12\textwidth]{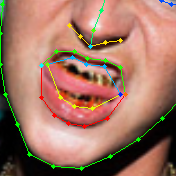} 
    \includegraphics[width=0.12\textwidth]{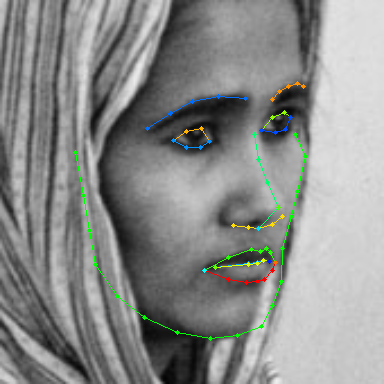} 
    \includegraphics[width=0.12\textwidth]{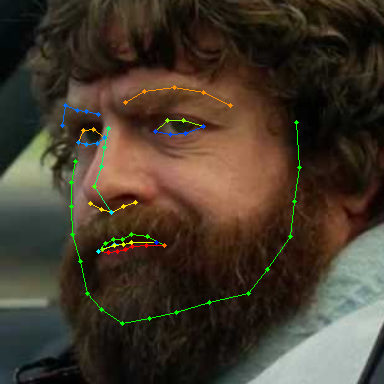} 
    \put(-491,3){\footnotesize \textcolor{white}{Lmk-line}} 
    \put(-429,3){\footnotesize \textcolor{white}{Lmk-line}} 
    \put(-367,3){\footnotesize \textcolor{white}{Lmk-line}} 
    \put(-305,3){\footnotesize \textcolor{white}{Lmk-line}} 
    \put(-243,3){\footnotesize \textcolor{white}{Lmk-line}} 
    \put(-181,3){\footnotesize \textcolor{white}{Lmk-line}} 
    \put(-119,3){\footnotesize \textcolor{white}{Lmk-line}} 
    \put( -57,3){\footnotesize \textcolor{white}{Lmk-line}} 
    \\
    \includegraphics[width=0.12\textwidth]{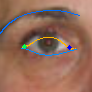} 
    \includegraphics[width=0.12\textwidth]{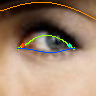} 
    \includegraphics[width=0.12\textwidth]{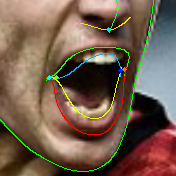} 
    \includegraphics[width=0.12\textwidth]{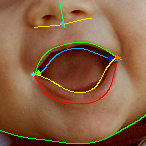} 
    \includegraphics[width=0.12\textwidth]{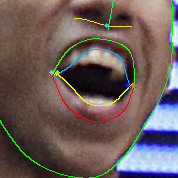} 
    \includegraphics[width=0.12\textwidth]{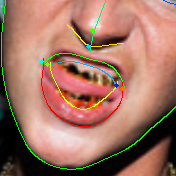} 
    \includegraphics[width=0.12\textwidth]{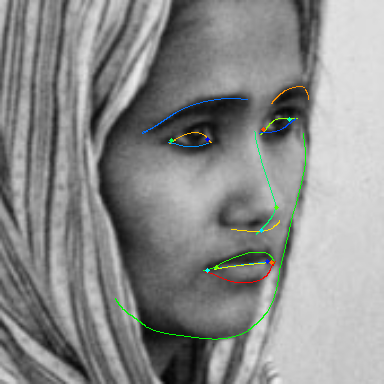} 
    \includegraphics[width=0.12\textwidth]{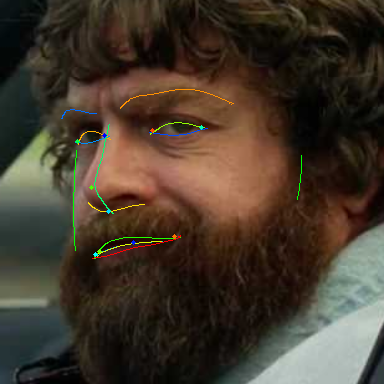} 
    \put(-491,3){\footnotesize \textcolor{white}{Line}} 
    \put(-429,3){\footnotesize \textcolor{white}{Line}} 
    \put(-367,3){\footnotesize \textcolor{white}{Line}} 
    \put(-305,3){\footnotesize \textcolor{white}{Line}} 
    \put(-243,3){\footnotesize \textcolor{white}{Line}} 
    \put(-181,3){\footnotesize \textcolor{white}{Line}} 
    \put(-119,3){\footnotesize \textcolor{white}{Line}} 
    \put( -57,3){\footnotesize \textcolor{white}{Line}} 
    \\
    \includegraphics[width=0.12\textwidth]{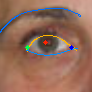} 
    \includegraphics[width=0.12\textwidth]{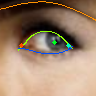} 
    \includegraphics[width=0.12\textwidth]{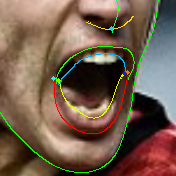} 
    \includegraphics[width=0.12\textwidth]{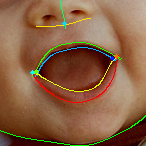} 
    \includegraphics[width=0.12\textwidth]{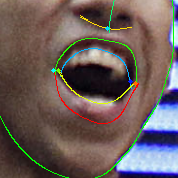} 
    \includegraphics[width=0.12\textwidth]{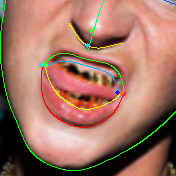} 
    \includegraphics[width=0.12\textwidth]{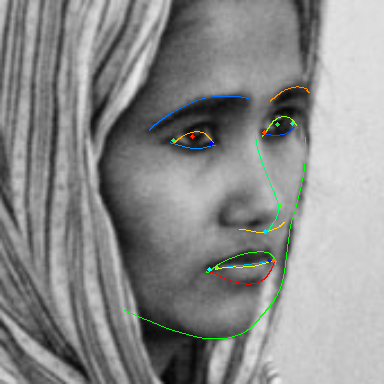} 
    \includegraphics[width=0.12\textwidth]{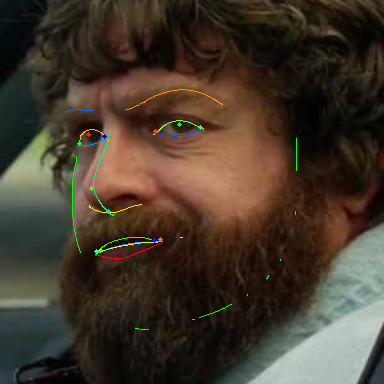}
    \put(-491,3){\footnotesize \textcolor{white}{ACE-Net}} 
    \put(-429,3){\footnotesize \textcolor{white}{ACE-Net}} 
    \put(-367,3){\footnotesize \textcolor{white}{ACE-Net}} 
    \put(-305,3){\footnotesize \textcolor{white}{ACE-Net}} 
    \put(-243,3){\footnotesize \textcolor{white}{ACE-Net}} 
    \put(-181,3){\footnotesize \textcolor{white}{ACE-Net}} 
    \put(-119,3){\footnotesize \textcolor{white}{ACE-Net}} 
    \put( -57,3){\footnotesize \textcolor{white}{ACE-Net}} 
    \\
    \includegraphics[width=0.12\textwidth]{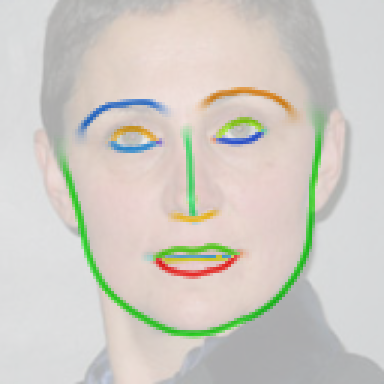} 
    \includegraphics[width=0.12\textwidth]{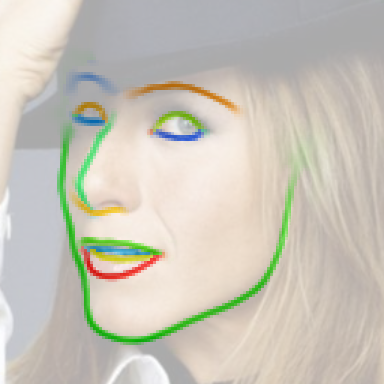} 
    \includegraphics[width=0.12\textwidth]{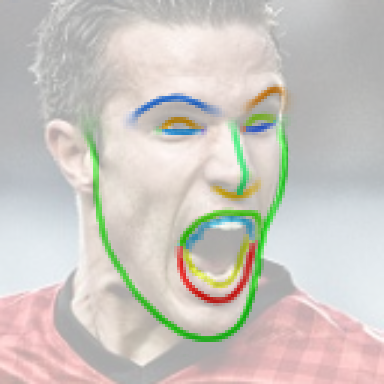} 
    \includegraphics[width=0.12\textwidth]{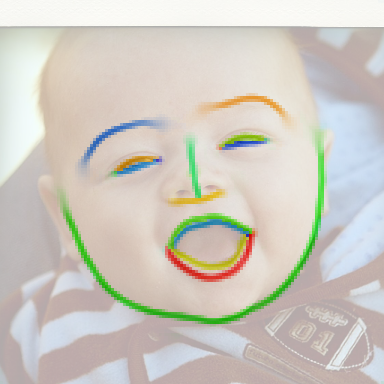} 
    \includegraphics[width=0.12\textwidth]{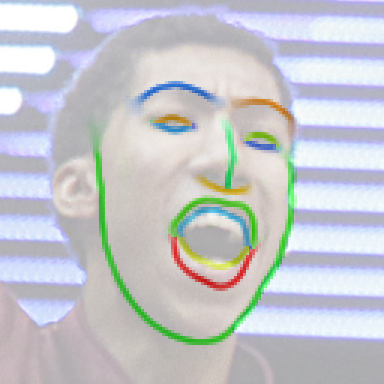} 
    \includegraphics[width=0.12\textwidth]{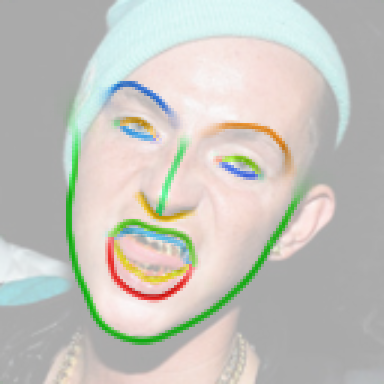} 
    \includegraphics[width=0.12\textwidth]{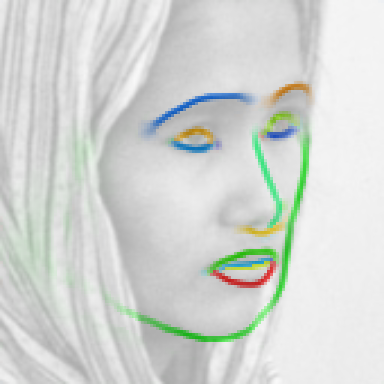} 
    \includegraphics[width=0.12\textwidth]{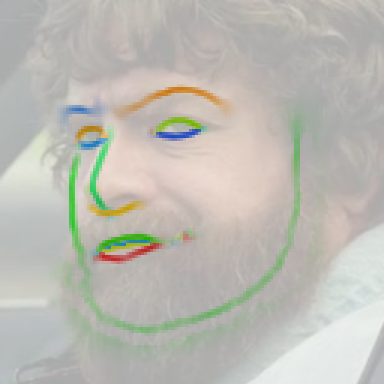}
    \put(-491,3){\footnotesize ACE-Net heat}
    \put(-429,3){\footnotesize ACE-Net heat}
    \put(-367,3){\footnotesize ACE-Net heat}
    \put(-305,3){\footnotesize ACE-Net heat}
    \put(-243,3){\footnotesize ACE-Net heat}
    \put(-181,3){\footnotesize ACE-Net heat}
    \put(-119,3){\footnotesize ACE-Net heat}
    \put( -57,3){\footnotesize ACE-Net heat}
    \caption{Qualitative results on 300-W testset. Each of the five rows shows (1) ground truth landmarks; (2) Lmk-line results; (3) Line results; (4) ACE-Net results; and (5) AC heatmaps predicted by ACE-Net. Due to space limitations, for some results we only show a small patch cropped using the bounding boxes visualized in the first row.}
    \label{fig:qualitative_examples}
\end{figure*}

To verify the robustness of \textbf{ACE-Net} against sparsity in training annotations, we reduce the number of landmarks used from 68 to only 36 (\figref{fig:dataset_landmarks}). We refer to our model trained with this reduced set of landmarks as \textbf{ACE-Net\textsuperscript{sparse}}. All its hyper-parameters, loss functions and the use of synthetic training data remains the same as the original ACE-Net. For fair comparison, we create models \textbf{GT36-line} and \textbf{GT36-interp} similarly to \textbf{GT68-line} and \textbf{GT68-interp} with the set of 36 landmarks.

\tabref{table:NME_HELEN_sparse} and \figref{fig:CED_HELEN_sparse} show their performance comparison. Although \textbf{ACE-Net\textsuperscript{sparse}} is trained with only 36 landmarks, its performance is very close to \textbf{ACE-Net} which is trained with 68 landmarks (NME\textsuperscript{AC} increase: $0.06$, AUC reduction: $0.58\%$). It clearly outperforms \textbf{GT36-line} and \textbf{GT36-interp}, both of which have a much larger performance drop compared with their \textbf{GT68} version (NME\textsuperscript{AC} increase: $0.94$ and $0.23$, AUC reduction: $9.43\%$ and $2.37\%$).

We also observe that \textbf{ACE-Net\textsuperscript{sparse}} outperforms \textbf{Lmk-line}, \textbf{GT68-line}, \textbf{GT68-interp}, \textbf{Line} and \textbf{Line$^*$}, despite them all trained with 68 landmarks. In particular, \textbf{ACE-Net\textsuperscript{sparse}} outperforms \textbf{Lmk-line} by a large margin (NME\textsuperscript{AC}: $1.38$ vs $2.20$, AUC: $86.19\%$ vs $77.95\%$). This comparison shows the ability of \textbf{ACE-Net} to make up for the reduced annotation density in the real training data.

\subsection{Qualitative Results}
\figref{fig:qualitative_examples} provides a qualitative comparison between \textbf{Lmk-line}, \textbf{Line} and \textbf{ACE-Net} models. It illustrates that landmarks and their corresponding line-contours predicted by \textbf{Lmk-line} are not accurate representations of true facial contours because the number of landmarks are limited. Contours predicted by \textbf{Line} appear more accurate, but they are really a smoothed version of line-contours and do not always follow local image cues, exhibiting similar error patterns as \textbf{Lmk-line}. In contrast, \textbf{ACE-Net} predicts facial contours whose shapes are more consistent with true facial contours, especially in places lacking training annotations such as inner lips. 

The last two columns in \figref{fig:qualitative_examples} show the most common failure case of AC based models. Our model is trained under the assumption that contours appear as ``ridges" in their corresponding heatmaps. When there is significant ambiguity, for example due to severe occlusion, predicted contour heatmaps are less ``peaky'' and get dropped during NMS+thresholding. However, in such cases fine-level alignment is usually out of the question anyway due to the lack of image cues for precise localization.

\section{Conclusion}
In this paper, we present \textbf{ACE-Net}, the first face alignment framework that learns to predict fine-level facial anchors and contours. Compared to contours generated blindly from landmarks or models trained with line-contours as ground truth, ACE-Net captures the fine-level details of facial contours with higher accuracy, especially in regions lacking annotations. Our approach allows the effective use of existing facial landmarks annotations as weak supervision through a novel ``contourness'' loss.
ACE-Net also demonstrates the potential of using synthetic data as a complement to sparsely labelled real data. Specifically, treating the sparsely labelled real data as weak supervision, synthetic data are able to help bridge the density gap and improve fine-level prediction accuracy.
We expect ACE-Net to make a significant impact on tasks dependent on accurate face alignment such as mesh fitting.

\begin{appendices}
\setcounter{equation}{14}

\section{Efficient Computation of Contourness $\mathcal{C}$}
In this section we derive the close-form solution of the optimization of $\mathcal{C}(H)$ in \eqref{eq:def_contourness_pixel} in the main paper:
\begin{align} \tag{6}
    \begin{split}
        \mathcal{C}_{\sigma}(H,p) = &-\min_{\theta} \sum_{i=-2\sigma}^{2\sigma} \sum_{j=-2\sigma}^{2\sigma} G_{\sigma}(i,j) \cdot \\
        &\norm{H_+(p_x+i, p_y+j) - T_{\sigma,\theta}(i, j)}^2
    \end{split}
    \label{eq:def_contourness_pixel}
\end{align}
where $H_+ = \max(0,H)$, and 
\begin{align}
    G_{\sigma}(i,j) &= e^{\dfrac{-(i^2+j^2)}{\sigma^2}}
    \label{eq:def_G} \\
    T_{\sigma,\theta}(i,j) &= \max\left(0, 1 - 2 \; \frac{(j\cos\theta - i\sin\theta)^2}{\sigma^2}\right)
    \label{eq:def_contourness_template} \tag{5}
\end{align}
We will omit $\sigma$ from subscripts since it is a constant hyper-parameter.

Since non-negative clipping $\max(0,\cdot)$ is already applied to the signal $H$ in \eqref{eq:def_contourness_pixel}, we drop the same clipping for the template $T$ in \eqref{eq:def_contourness_template} which does not change the value of optimal $\mathcal{C}(H)$. Thus, the definition of $T$ becomes
\begin{equation}
    T_{\theta}(i,j) = 1 - 2 \; \frac{(j\cos\theta - i\sin\theta)^2}{\sigma^2}
    \label{eq:def_contourness_template_simplified}
\end{equation}

Let $\Delta p = (i,j)$. By expanding $\norm{\cdot}^2$ in \eqref{eq:def_contourness_pixel} we get
\begin{align}
    \begin{split}
        \mathcal{C}(H,p) = -\min_{\theta} &\sum_{\Delta p} G(\Delta p) \cdot H_+(p + \Delta p)^2 + \\
        &\sum_{\Delta p} G(\Delta p) \cdot \left( -2H_+(p + \Delta p) T_{\theta}(\Delta p)\right) + \\
        &\sum_{\Delta p} G(\Delta p) \cdot T_{\theta}(\Delta p)^2
    \end{split}
    \label{eq:def_contourness_pixel_expanded}
\end{align}
Here the last term $\sum_{\Delta p} G(\Delta p) \cdot T_{\theta}(\Delta p)^2$ is a constant number given $\sigma$. Therefore, we drop it as it does not affect the optimization.

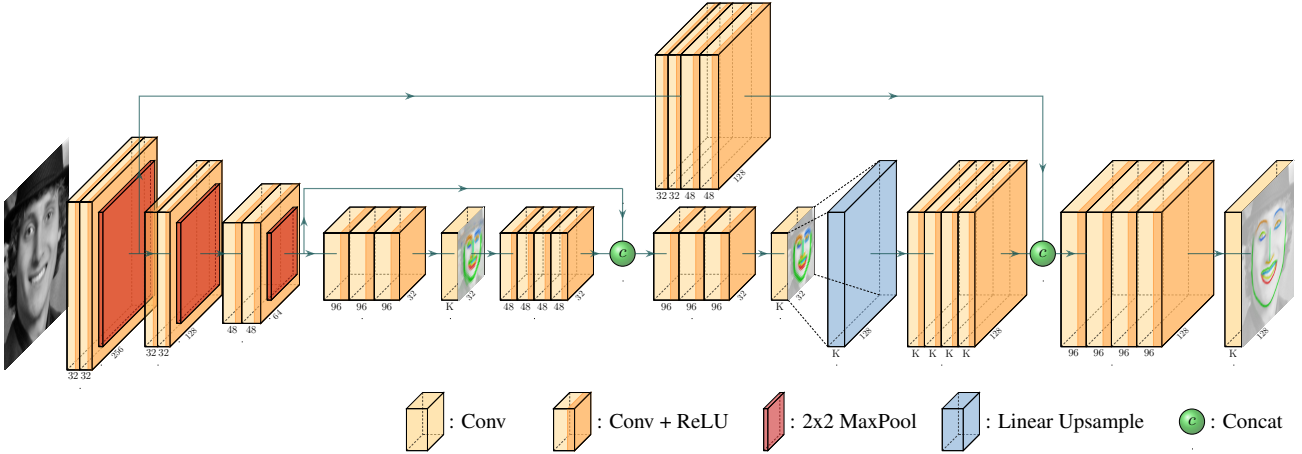
\begin{figure*}[!t]
    \includegraphics[width=\textwidth]{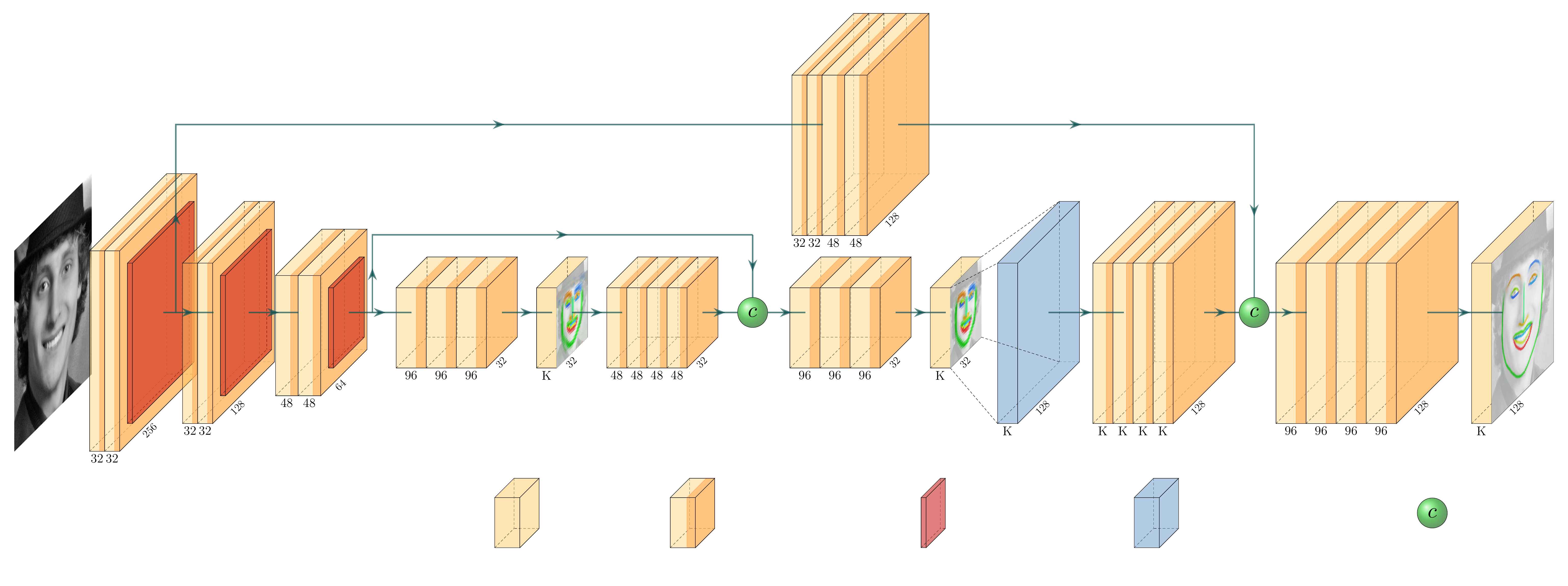}
    \put(-324,20.5){\footnotesize : Conv}
    \put(-268,20.5){\footnotesize : Conv + ReLU}
    \put(-195,20.5){\footnotesize : 2x2 MaxPool}
    \put(-121,20.5){\footnotesize : Linear Upsample}
    \put(-36,20.5){\footnotesize : Concat}
    \vspace{-1em}
    \caption{Network architecture of our example ACE-Net model with all filter sizes and layer sizes annotated. Zoom in for a clearer view. K is the number of heatmaps for each image. Our example architecture takes grayscale images as input and outputs 3 sets of AC heatmaps (the output layers are visualized with heatmaps overlay). The first two sets of heatmaps are only used for intermediate supervision at training time, and are ignored at inference time.}
    \label{fig:arch}
\end{figure*}

To efficiently compute contourness $\mathcal{C}(H)$ for every pixel $p$ in $H$, we rewrite \eqref{eq:def_contourness_pixel_expanded} with convolution:
\begin{align}
    \begin{split}
        \mathcal{C}(H) &= -\min_{\theta} \left(H_+^2 \otimes G - 2H_+ \otimes (T_{\theta} \cdot G)\right) \\
                       &= - H_+^2 \otimes G - \min_{\theta} \left(-2H_+ \otimes (T_{\theta} \cdot G)\right) \\
                       &= - H_+^2 \otimes G + 2\max_{\theta} \left(H_+ \otimes (T_{\theta} \cdot G)\right)
    \end{split}
    \label{eq:def_contourness_pixel_conv}
\end{align}
Note that $T_{\theta} \cdot G$ is a steerable filter (similar to $G_2^{\theta}$ in Table III from \cite{freeman1991design}). According to \cite{freeman1991design}, $H_+ \otimes (T_{\theta} \cdot G)$ can be computed as
\begin{align}
    \begin{split}
        H_+ \otimes (T_{\theta} \cdot G) =\; &\left( H_+ \otimes G_{2a} \right) \cdot \sin^2\theta \\
                                             -&\left( H_+ \otimes G_{2b} \right) \cdot 2\cos\theta\sin\theta \\
                                             +&\left( H_+ \otimes G_{2c} \right) \cdot \cos^2\theta 
        \label{eq:compute_HTG_part1}
    \end{split}
\end{align}
where $G_{2a}$, $G_{2a}$ and $G_{2a}$ are convolutional filters:
\begin{align}
    \begin{split}
        G_{2a}(x,y) &= \left(1 - \dfrac{2x^2}{\sigma^2}\right) e^{-\frac{x^2+y^2}{\sigma^2}}\\
        G_{2b}(x,y) &= -\dfrac{2xy}{\sigma^2} e^{-\frac{x^2+y^2}{\sigma^2}}\\
        G_{2c}(x,y) &= \left(1 - \dfrac{2y^2}{\sigma^2}\right) e^{-\frac{x^2+y^2}{\sigma^2}}
        \label{eq:def_contourness_filters}
    \end{split}
\end{align}
(Our filters and coefficients definition here are slightly different from Table III in \cite{freeman1991design}, because $\theta$ in \cite{freeman1991design} is defined as the normal instead of the contour orientation.) \\
Now, let $R_a$, $R_b$, $R_c$ be the convolution responses:
\begin{align}
    \begin{split}
        R_a &= H_+ \otimes G_{2a}\\
        R_b &= H_+ \otimes G_{2b}\\
        R_c &= H_+ \otimes G_{2c}
        \label{eq:def_contourness_filters}
    \end{split}
\end{align}
then \eqref{eq:compute_HTG_part1} becomes
\begin{align}
    \begin{split}
        H_+ \otimes \left(T_{\theta} \cdot G\right) &= R_a\sin^2\theta  - R_b \cdot 2\cos\theta\sin\theta + R_c\cos^2\theta \\
        &= R_a\dfrac{1 - \cos2\theta}{2} - R_b\sin2\theta + R_c\dfrac{1 + \cos2\theta}{2} \\
        &= \dfrac{R_c + R_a}{2} + \dfrac{R_c - R_a}{2}\cos2\theta - R_b\sin2\theta \\
        &= \dfrac{R_c + R_a}{2} + \\
           &\hskip 3.5em \cos(2\theta - \phi) \sqrt{\left(\dfrac{R_c - R_a}{2}\right)^2 + R_b^2}
        \label{eq:compute_HTG_part2}
    \end{split}
\end{align}
Thus we have
\begin{equation}
    \max_{\theta} \left(H_+ \otimes \left(T_{\theta} \cdot G\right)\right) = \dfrac{R_c + R_a}{2} + \sqrt{\left(\dfrac{R_c - R_a}{2}\right)^2 + R_b^2}
    \label{eq:compute_max_HTG}
\end{equation}
Plugging \eqref{eq:compute_max_HTG} back to \eqref{eq:def_contourness_pixel_conv}, we get the closed-form solution of $\mathcal{C}(H)$:
\begin{equation}
    \mathcal{C}(H) = R_a + R_c + \sqrt{(R_a - R_c)^2 + 4R_b^2} - H_+^2 \otimes G
    \label{eq:def_contourness}
\end{equation}
\eqref{eq:def_contourness} shows that the contourness map $\mathcal{C}(H)$ can be very efficiently computed simply with convolutions, and therefore can be integrated into loss functions.

From \eqref{eq:compute_HTG_part2}, we also derive the closed-form solution of the contour orientation map $\mathcal{O}(H)$:
\begin{align}
    \begin{split}
        \mathcal{O}(H) &= \argmax_{\theta} \left(H_+ \otimes \left(T_{\theta} \cdot G\right)\right) \\
                       &= \frac{\phi}{2} \\
                       &= \frac{\arctan(-2R_b, \; R_c - R_a)}{2}
        \label{eq:def_contour_orientation}
    \end{split}
\end{align}
and the contour normal map $\mathcal{N}(H)$ used in our AC extraction module is computed as
\begin{align}
    \begin{split}
        \mathcal{N}(H) = \frac{\pi}{2} + \mathcal{O}(H) = \frac{\arctan(2R_b, \; R_a - R_c)}{2}
        \label{eq:def_contour_normal}
    \end{split}
\end{align}

\section{Example Network Architecture}
\figref{fig:arch} fully illustrates the architecture of our example ACE-Net model used for our experiments. We predict $N_A = 12$ anchors and $N_C = 13$ contours per image (i.e., $K = 25$ heatmaps per image). The names of anchors and contours are listed in \tabref{table:AC_names}. Our example architecture is picked due to its simplicity (inference takes around 17ms per image on a 8GB GeForce GTX 1080 GPU). Since our model has intermediate outputs, we apply losses to intermediate outputs as well for better intermediate supervision. Because the intermediate outputs have very low resolution equal to 1/8 of the input, we just treat line-contours as ground truth and apply fully supervised loss. This intermediate loss is not required, especially when a network architecture without intermediate outputs (such as Stacked Hourglass) is chosen instead. ACE-Net does not rely on any specific network architecture.

\begin{table}[!tbp]
    \centering
    \begin{tabular}{ c|c } 
        \hline
        Anchors & Contours \\ 
        \hline
        right eye inner corner & right eyebrow center-line \\ 
        right eye outer corner & left eyebrow center-line \\ 
        left eye inner corner & right eye upper lid \\ 
        left eye outer corner & right eye lower lid \\ 
        right iris center & left eye upper lid \\ 
        left iris center & left eye lower lid \\ 
        nose tip & nose ridge \\ 
        nose bottom center & nose bottom boundary \\ 
        mouth right outer corner & mouth upper lip outer \\ 
        mouth left outer corner & mouth lower lip outer \\ 
        mouth right inner corner & mouth upper lip inner \\ 
        mouth left inner corner & mouth lower lip inner \\ 
         & chin boundary \\ 
        \hline
    \end{tabular}
    \caption{List of facial anchors and contours. The definitions of anchors and contours are illustrated by figures in the main paper.}
    \label{table:AC_names}
\end{table}

\end{appendices}

{\small
\bibliographystyle{ieee_fullname}
\bibliography{arxiv}
}

\end{document}